\documentclass{article} 
\usepackage{iclr2026_conference,times}

\usepackage{amsmath,amsfonts,bm}

\def\eqref#1{equation~\ref{#1}}

\def\1{\bm{1}}

\DeclareMathAlphabet{\mathsfit}{\encodingdefault}{\sfdefault}{m}{sl}
\SetMathAlphabet{\mathsfit}{bold}{\encodingdefault}{\sfdefault}{bx}{n}

\usepackage{hyperref}
\usepackage{url}

\usepackage{booktabs}       
\usepackage{graphicx}
\usepackage{multirow}
\usepackage{amsmath}
\usepackage{wrapfig}
\usepackage{array}
\usepackage{caption}

\title{Adaptive Budget Allocation\\ for Orthogonal-Subspace Adapter Tuning\\ in LLMs Continual Learning}

\author{
  Zhiyi Wan\textsuperscript{1}, 
  Wanrou Du\textsuperscript{1},
  Liang Li\textsuperscript{2},
  Miao Pan\textsuperscript{3},
  Xiaoqi Qin\textsuperscript{1} \\
  \textsuperscript{1}Beijing University of Posts and Telecommunications, \textsuperscript{2}Pengcheng Laboratory, \textsuperscript{3}University of Houston \\
  \texttt{\{wzy10, wanroudu, xiaoqiqin\}@bupt.edu.cn}, 
  \texttt{lil03@pcl.ac.cn},
  \texttt{mpan2@uh.edu}
}

\iclrfinalcopy 

\begin{document}

\maketitle

\begin{abstract}
Large language models (LLMs) often suffer from catastrophic forgetting in continual learning (CL) scenarios, where performance on previously learned tasks degrades severely while training on sequentially arriving tasks. 
Although pioneering CL approaches using orthogonal subspaces can mitigate task interference, they typically employ fixed budget allocation, neglecting the varying complexity across tasks and layers. 
Besides, recent budget-adaptive tuning methods for LLMs often adopt multi-stage paradigms that decouple optimization and budget allocation. Such decoupling results in potential misalignment, which hinders those approaches' practical application in CL scenarios. 
To address these limitations, we propose OA-Adapter, a novel parameter-efficient approach for continual learning in LLMs that unifies dynamic budget adaptation with orthogonal subspace learning in an end-to-end training stage. 
Specifically, OA-Adapter introduces a dynamic bottleneck dimension adaptation mechanism that simultaneously allocates an efficient parameter budget and optimizes task objectives without misalignment.
To effectively preserve previously acquired knowledge while coordinating with the dynamic budget allocation, orthogonal constraints are applied specifically between the parameter subspace of the current task and the dynamically allocated parameter subspaces of historical tasks. 
Experimental results on continual learning benchmarks demonstrate that OA-Adapter outperforms state-of-the-art methods in both accuracy and parameter efficiency. OA-Adapter achieves higher average accuracy while using \(58.5\%\) fewer parameters on the standard CL benchmark, and maintains its advantages on two larger benchmarks comprising 15 tasks. 
\end{abstract}

\section{Introduction}
Recent advances in large language models (LLMs) have transformed artificial intelligence by demonstrating remarkable capabilities across diverse domains~\citep{DBLP:conf/ijcai/0001SNYZ24,DBLP:journals/corr/abs-2302-13971}. 
However, real‐world deployment demands that LLMs continually adapt to evolving user needs and emerging tasks while retaining previously acquired knowledge—a prerequisite for sustainable lifelong learning~\citep{DBLP:journals/chinaf/XiCGHDHZWJZZFWXZWJZLYDW25,DBLP:journals/corr/abs-2310-06762}.

Parameter-efficient fine-tuning (PEFT) methods, such as adapter modules~\citep{DBLP:conf/icml/HoulsbyGJMLGAG19} and low-rank adaptation (LoRA)~\citep{DBLP:conf/iclr/HuSWALWWC22}, enable task‐specific adaptation by updating only \(0.01\%-4\%\) of model weights~\citep{DBLP:journals/corr/abs-2404-13506}). 
While originally designed to reduce computational costs for single‐task tuning~\citep{DBLP:conf/acl/Zhou0KZGP24,DBLP:conf/nlpcc/ChenLWL24}, these methods struggle in continual learning with sequentially arriving tasks. 
Sequentially tuning to arriving tasks induces catastrophic forgetting‐severe degradation of performance on prior tasks~\citep{mccloskey1989catastrophic}. 
One intuitive solution is to store task-specific adapters for each new task. However, this approach consumes substantial storage resources and leads to considerable inflexibility in multi-task deployments. Alternatively, retraining models with archived historical and additional new data necessitates frequent model updates and large data repositories~\citep{DBLP:journals/corr/abs-2404-16789}. Both strategies are prohibitively costly and impractical in resource-constrained environments.

To efficiently adapt LLMs to downstream tasks while preserving previously acquired knowledge, researchers proposed continual learning (CL) methods. However, most CL approaches operate within shared parameter spaces across tasks, which inherently induces cross-task interference~\citep{DBLP:journals/pami/WangZSZ24, DBLP:journals/corr/abs-2404-16789, chaudhry2019tiny, DBLP:conf/nips/ShiW23, DBLP:conf/cvpr/RebuffiKSL17, DBLP:conf/eccv/AljundiBERT18,DBLP:conf/icml/Schwarz0LGTPH18,rongali2020continual,DBLP:conf/cvpr/LinCL22}. 
Furthermore, unlike conventional CL (e.g., incremental image classification) that typically handle tasks with limited distribution shifts , CL of LLMs often deals with substantially divergent task distributions, thus significantly amplifying interference when tuning in a shared parameter space. 
Some architecture-based methods further construct task-specific parameters to resolve this problem, but they heavily rely on explicit task ID during inference~\citep{DBLP:conf/iclr/RazdaibiedinaMH23,DBLP:conf/iclr/JangYYSHKCS22,DBLP:conf/naacl/JinZZ00WA022,DBLP:journals/corr/abs-2401-03129,DBLP:conf/bibm/YanXSYLR23}. 
Recent researches in orthogonal subspace learning~\citep{DBLP:conf/aistats/FarajtabarAML20,DBLP:conf/aaai/GuoH0L22,DBLP:conf/emnlp/WangCGXBZZGH23} offer a promising alternative by restricting task-specific updates to mutually orthogonal parameter subspaces to eliminate interference, without task ID dependency during inference. 
However, existing methods typically rely on a fixed budget allocation, assigning the same subspace dimensionality to different task and layer. This rigid strategy overlooks the heterogeneity of task complexity and layer-specific adaptation needs, leading to inefficient parameter utilization, allocating excessive resource to simple tasks while under-allocating resources to more complex ones. Such inflexible allocation hinder LLMs' continuous adaptation capabilities in practice. 

To achieve dynamic budget allocation, emerging budget-adaptive PEFT methods like AdaLoRA~\citep{zhang2023adaloraadaptivebudgetallocation}, ElaLoRA~\citep{chang2025elalora}, and DiffoRA~\citep{DBLP:journals/corr/abs-2502-08905} proposed multi-stage paradigms with sequential optimization and budget adjustment phases. 
Such decoupled optimization may create misalignment between fine-tuning objectives and budget allocation. Moreover, the inherent complexity of multi-stage designs introduces substantial computational overhead and engineering challenges, limiting their practicality for continual learning systems. 

We propose OA-Adapter to address these issues, a novel parameter-efficient approach for continual learning in LLMs. Instead of manually assigning a fixed budget, OA-Adapter automatically adjusts the parameter budget for each task and layer based on the task difficulty and model capacity. 

Our key contributions are as follows:
\begin{itemize}
    \item We propose OA-Adapter, a novel parameter-efficient approach for continual learning in LLMs that unifies dynamic budget adaptation with orthogonal subspace learning in an end-to-end training stage. To the best of our knowledge, this is the first work to integrate budget adaptation into parameter-efficient fine-tuning for continual learning in LLMs.
    \item We design a dynamic bottleneck dimension adaptation mechanism that simultaneously allocates an efficient parameter budget and optimizes task objectives without misalignment.
    \item We demonstrate that the parameter budget requirements in continual learning vary with task difficulty and position in the training sequence, and that the budget requirements of parameters in different layers vary, which emphasizes the necessity of our approach.
    \item Experimental results demonstrate that OA-Adapter outperforms state-of-the-art methods in both accuracy and parameter efficiency. OA-Adapter achieves higher average accuracy with \(58.5\%\) fewer parameters on the standard CL benchmark and maintains its advantages on two larger benchmarks comprising 15 tasks. 
\end{itemize}

\section{Related Work.}
\label{Related work}
\paragraph{Continual Learning for LLMs.}
Existing continual learning (CL) techniques for LLMs are mainly based on replay, regularization, and architecture modification, which have been most extensively applied in this context~\citep{DBLP:journals/pami/WangZSZ24,DBLP:journals/corr/abs-2404-16789}.
Replay-based methods~\citep{chaudhry2019tiny,DBLP:conf/nips/ShiW23,DBLP:conf/cvpr/RebuffiKSL17} store a subset of past data to rehearse on previous tasks during new training. However, this approach raises significant privacy issues and imposes high storage and computational costs, especially for LLMs.
Regularization-based methods~\citep{DBLP:conf/eccv/AljundiBERT18,DBLP:conf/icml/Schwarz0LGTPH18,rongali2020continual,DBLP:conf/cvpr/LinCL22} introduce a regularization term that penalizes significant weight deviations across tasks. However, their computational complexity grows rapidly with the number of tasks, and performance on previous tasks deteriorates significantly.
These above approaches primarily focus on learning all incremental tasks within a shared parameter space, which is a major contributor to task interference. In contrast, many architecture-based methods~\citep{DBLP:conf/iclr/RazdaibiedinaMH23,DBLP:conf/iclr/JangYYSHKCS22,DBLP:conf/naacl/JinZZ00WA022,DBLP:journals/corr/abs-2401-03129,DBLP:conf/bibm/YanXSYLR23} address this challenge by incorporating task-specific components, isolated parameters, or dedicated pathways within the model. They essentially learn task-specific expert modules for different tasks, so rely heavily on explicit task ID during inference. 
Recent advancements have explored a promising direction that can retain generalization capacity while reducing task interference without requiring explicit task ID. These methods~\citep{DBLP:conf/aistats/FarajtabarAML20, DBLP:conf/aaai/GuoH0L22, DBLP:conf/emnlp/WangCGXBZZGH23} constrain weight updates for each task to lie within mutually orthogonal subspaces of the high-dimensional parameter space. By doing so, they effectively decouple the optimization directions of different tasks, thereby mitigating task interference. However, a common limitation of existing orthogonal subspace methods is the fixed parameter budget for all tasks and layers. 
\paragraph{Adaptive Budget Tuning.}
Existing budget-adaptive tuning methods for LLMs commonly adopt a multi-stage design where budget adaptation is a separate stage from optimization. 
For example, DiffoRA~\citep{DBLP:journals/corr/abs-2502-08905} trains a LoRA module for each layer in the first stage, and then selects a subset of these modules based on the "Difference-aware Adaptive Matrix" in a subsequent stage. ElaLoRA~\citep{chang2025elalora} involves a "Warm-up" fine-tuning stage, followed by a "Dynamic Rank Adjustment" stage to adapt LoRA module ranks, and finally a "Stabilization" fine-tuning stage. 
AdaLoRA~\citep{zhang2023adaloraadaptivebudgetallocation} injects rank pruning operations after iterations by applying SVD to selectively retain low-rank components. 
While these methods achieve strong performance on single-task datasets, their multi-stage design introduces substantial computational and engineering overhead that compromises practicality and scalability, while creating misalignment between optimization objectives and budget adaptation criteria that ultimately hinders the achievement of truly optimal solutions.
And their design is restricted to unidirectional rank reduction, which does not allow for bidirectional budget adaptation. These limitations pose significant challenges when extending such methods to continual learning, which demands efficient, unified, and adaptive training processes for sequential tasks. 
Moreover, although recent methods~\citep{liu-etal-2024-alora, chang2025elalora,ding-etal-2023-sparse} provide strong interpretability, they rely on task-specific heuristic configurations rather than adapting to the inherent characteristics of sequentially arriving tasks.

\section{Methodology}
\label{Methodology}
\begin{figure}[!t] 
\centering
\includegraphics[width=0.95 \linewidth]{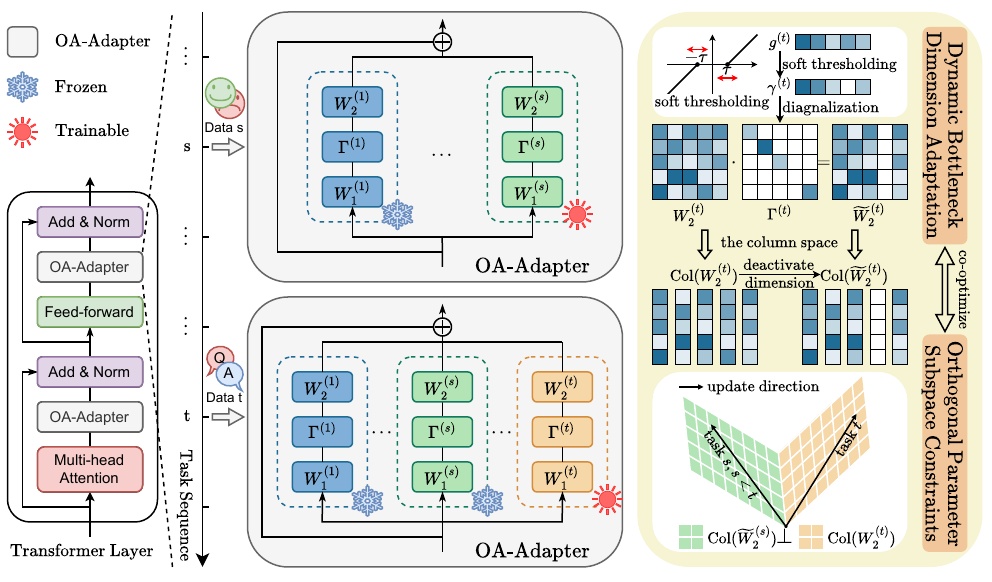}
\caption{The OA-Adapter framework for LLM continual learning. Each task-specific OA-Adapter module (task \(t\)) comprises three core components: (1) a down-projection layer \(\mathcal{W}_1^{(t)}\), (2) a  diagonal mask \(\Gamma^{(t)}\) with trainable threshold \(\tau^{(t)}\), (3) and an up-projection layer \(\mathcal{W}_2^{(t)}\). 
The dynamic masking mechanism enables bidirectional dimension adaptation through activation/deactivation of latent dimensions. Orthogonal subspace constraints are enforced between the column space of the \(t\)-th task parameters \(\mathrm{Col}(\mathcal{W}_2^{(t)})\) and the dynamically allocated parameter subspaces of historical tasks \(\mathrm{Col}(\widetilde{\mathcal{W}}_2^{(s)})\) (for \(\ s < t\)). Here, \(\widetilde{\mathcal{W}}_2^{(s)}\) incorporates only the activated dimensions from the \(s\)-th task.}
\label{fig:Model_Structure}
\end{figure}

In this section, we introduce OA-Adapter, a novel framework for continual learning in LLMs that simultaneously improves parameter efficiency and mitigates catastrophic forgetting in an end-to-end training stage, as illustrated in Figure~\ref{fig:Model_Structure}. We first describe its architectural design, including core components and computation flow. Then, we analyze the mathematical foundations of the dynamic bottleneck dimension adaptation, demonstrating how trainable thresholds enable bidirectional activation and deactivation of dimensions during an end-to-end training phase. Finally, we formalize the orthogonal parameter subspace constraints mechanism and explain how it works in concert with the dynamic bottleneck dimension adaptation to achieve parameter-efficient continual learning.

\subsection{Module Structure}

\paragraph{Standard Adapter.}
To enable efficient adaptation of pre-trained language models (PLMs) to downstream tasks, adapter modules inject lightweight trainable parameters into PLMs while keeping the original weights frozen. These modules employ a bottleneck architecture to minimize trainable parameters, consisting of a down-projection layer that reduces the original \(d\)-dimensional representation \(x\) to a lower dimension \(r\), a nonlinear activation function \(f(\cdot)\), and an up-projection layer that restores the features to dimension \(d\). The architecture ensures near-zero initialization of the projection layers while maintaining a skip connection to preserve the original features during initial training stages. For an input representation \(x \in \mathbb{R}^{d}\), the output \(y \in \mathbb{R}^{d}\) can be formalized as:
\begin{equation}
y=x + \mathcal{W}_2 \cdot f(\mathcal{W}_1 \cdot x + b_1) + b_2,
\end{equation}
where \(\mathcal{W}_1 \in \mathbb{R}^{r \times d}, \mathcal{W}_2 \in \mathbb{R}^{d \times r}\) denote the down-projection and up-projection matrices. \( \mathbf{b}_1 \in \mathbb{R}^r \) and \( \mathbf{b}_2 \in \mathbb{R}^d \) denote the bias terms, respectively, with bottleneck dimension \( r \ll d \).

\paragraph{OA-Adapter.}
Building upon the standard Adapter’s bottleneck architecture, OA-Adapter introduces two structural modifications: 1) the removal of bias terms in projection layers to create a bias-free parameter space containing only linear transformations, and 2) the replacement of static non-linear activations with a trainable diagonal masking matrix \(\Gamma\) that dynamically adjusts the bottleneck dimension, as detailed in Section~\ref{Section:Dynamic Bottleneck Dimension Adaptation}. These modifications enable the enforcement of orthogonal parameter subspace constraints, as detailed in Section~\ref{Section:Orthogonal Parameter Subspace Constraints for Continual Learning}, to mitigate cross-task interference and co-optimization of budget adaptation with continual learning in an end-to-end training phase. Specifically, the forward computation of OA-Adapter operates as follows:
\begin{equation}
y=x + \mathcal{W}_2 \cdot \Gamma \cdot \mathcal{W}_1 \cdot x,
\end{equation}
where \(\mathcal{W}_1 \in \mathbb{R}^{r_{\max} \times d}\) and \(\mathcal{W}_2 \in \mathbb{R}^{d \times r_{\max}}\) denote the down-projection and up-projection matrices, respectively, and \(\Gamma \in \mathbb{R}^{r_{\max} \times r_{\max}}\) is a trainable diagonal masking matrix. Here, \(r_{\max} \ll d\) represents the pre-defined maximum bottleneck dimension.

\subsection{Dynamic Bottleneck Dimension Adaptation}
\label{Section:Dynamic Bottleneck Dimension Adaptation}

\paragraph{Adaptation Mechanism.}
To dynamically allocate parameter budget, we adjust the effective bottleneck dimensions of OA-Adapter using a trainable diagonal masking matrix \(\Gamma\in \mathbb{R}^{r_{\max} \times r_{\max}}: \Gamma= \mathrm{diag}(\gamma)\). The sparsity of the vector \(\gamma \in \mathbb{R}^{r_{\max}}\) is controlled via a soft thresholding mechanism applied to a trainable vector \(g \in \mathbb{R}^{r_{\max}}\). Specifically, each diagonal entry \(\gamma_i\) is computed as:
\begin{equation}
\gamma_i = \mathrm{soft}(g_i; \tau)= \mathrm{sign}(g_i) \cdot \max(|g_i| - \tau, 0),
\end{equation}
where \(\tau > 0\) is a trainable threshold that dynamically modulates the sparsity level of \(\Gamma\) throughout the training process. The projection path of OA-Adapter can then be equivalently reformulated as:
\begin{equation}
\mathcal{W}_2 \cdot \Gamma \cdot \mathcal{W}_1 = \sum_{i=1}^{r_{\max}} \gamma_i \cdot {\mathcal{W}_2[:,i] \otimes \mathcal{W}_1[i,:]},
\end{equation}
where \(\otimes\) denotes the outer product. This decomposition clearly demonstrates how each \(\gamma_i\) dynamically adjusts the contribution of the \(i^\mathrm{th}\) latent dimension pair: when \(|g_i| \leq \tau\), we have \(\gamma_i=0\), causing both the \(i^\mathrm{th}\) column of \(\mathcal{W}_2\) and \(i^\mathrm{th}\) row of \(\mathcal{W}_1\) to be disabled, effectively deactivating the corresponding dimension. This mechanism adaptively controls the bottleneck dimension through \(r_{\mathrm{eff}} = \|\gamma\|_0\), where \(\|\gamma\|_0\) represents the count of non-zero entries in \(\gamma\).

\paragraph{Gradient Analysis.}
Our method’s bidirectional dimension adaptation capability, enabled by the trainable threshold \(\tau\), offers critical advantages. When \(|g_i| \leq \tau\), the corresponding dimension pair is deactivated by setting \(\gamma_i=0\). This zeros the \(i\)-th diagonal entry of the masking matrix \(\Gamma\), effectively removing that dimension’s contribution in forward propagation. However, this operation also blocks gradient flow from the downstream loss \(\mathcal{L}\) to \(g_i\) when \(\gamma_i=0\). To clarify why \(g_i\) becomes non-trainable in such cases, consider the gradient calculation via chain rule:
\begin{equation}
\frac{\partial \mathcal{L}}{\partial g_i} = \frac{\partial \mathcal{L}}{\partial \gamma_i} \frac{\partial \gamma_i}{\partial g_i}.
\label{loss_L_gi}
\end{equation}
The derivative \(\frac{\partial \gamma_i}{\partial g_i}\) of the soft thresholding function equals \(\mathrm{sign}(g_i)\) when \(|g_i| > \tau\), but critically becomes \(0\) when \(|g_i| \leq \tau\). This implies that deactivated dimensions (where \(|g_i| \leq \tau\)) produce zero gradients in Equation (\ref{loss_L_gi}) due to \(\frac{\partial \gamma_i}{\partial g_i}=0\), blocking gradient updates through \(g_i\). With a fixed threshold \(\tau\), such dimensions would remain permanently disabled throughout training. Crucially, our method implements \(\tau\) as a learnable parameter shared across all dimensions within each OA-Adapter module. Thus, the gradient of \(\tau\) with respect to the total loss \(\mathcal{L}\) is:
\begin{equation}
\frac{\partial \mathcal{L}}{\partial \tau} = \sum_{i=1}^{r_{\max}} \frac{\partial \mathcal{L}}{\partial \gamma_i} \frac{\partial \gamma_i}{\partial \tau},
\end{equation}
where \(\frac{\partial \gamma_i}{\partial \tau} = -\mathrm{sign}(g_i)\) when \(|g_i| > \tau\) and 0 otherwise. This derivative relationship ensures threshold updates are primarily governed by dimensions exceeding the current \(\tau\). As \(\tau\) evolves during training, dimensions previously deactivated with \(|g_i| \leq \tau\) may become reactivated when they satisfy \(|g_i| > \tau\) under the updated threshold, thereby reactivating their corresponding projection paths. This bidirectional adaptation mechanism automatically suppresses dimensions while maintaining their potential for reactivation in later training iterations. 
The bidirectional nature of this dynamic parameter budget adaptation approach ensures optimal parameter allocation that continuously adapts to the evolving requirements of sequential tasks in continual learning. 

\subsection{Orthogonal Parameter Subspace Constraints for Continual Learning}
\label{Section:Orthogonal Parameter Subspace Constraints for Continual Learning}

\paragraph{Continual Learning Setup.}
Continual learning focuses on incrementally acquiring knowledge from evolving data distributions of sequential tasks while mitigating catastrophic forgetting of previously acquired knowledge. Formally, models are trained on a sequential stream of tasks denoted as \{\(D_1, D_2,\cdots, D_t\)\}. Each task \(D_t={(x_t^i,y_t^i)}_{i=1}^{n_t}\) consists of input instances \(x_t^i \in \mathcal{X}_t\) paired with corresponding labels \(y_t^i \in \mathcal{Y}_t\), where \(\mathcal{X}_t\) and \(\mathcal{Y}_t\) represent the task-specific input and label spaces. During the training phase for task \(D_t\), model parameters \(\Phi\) are updated exclusively using data from \(D_t\). The objective of continual learning can be formalized as optimizing:
\begin{equation}
\max_{\Phi} \sum_{t=1}^T \sum_{\{x_i^t, y_i^t\} \in \mathcal{D}_t} \log P_\Phi(y_t^i \mid x_t^i).
\end{equation}
\paragraph{Orthogonal Parameter Subspace Constraints.}
Catastrophic forgetting arises when subsequent adaptations overwrite parameters critical for previous tasks. To mitigate this, we introduce orthogonality constraints that enforce parameter updates across tasks to occupy mutually independent subspaces. Let \(\Delta \Phi_k\) represent the OA-Adapter's parameter for the \(k\)-th task. We have:
\begin{equation}
\Delta \Phi_k 
= \mathcal{W}_2^{(k)} \cdot \Gamma^{(k)} \cdot \mathcal{W}_1^{(k)}
= {\widetilde{\mathcal{W}}_2^{(k)}} \cdot \mathcal{W}_1^{(k)}
\label{parameter}
\end{equation}
Here, the columns of \(\widetilde{\mathcal{W}}_2^{(k)}\) serve as orthogonal basis vectors spanning the parameter update subspace for the \(k\)-th task, while \(\mathcal{W}_1^{(k)}\) determines how these basis vectors are combined. We therefore formally define the task-specific parameter subspace as the column space of \(\widetilde{\mathcal{W}}_2^{(k)}\), which intrinsically aligns with the activated dimensions for the \(k\)-th task through the dimension-selective masking operation of \(\Gamma^{(k)}\).
Thus, we enforce strict orthogonality to new OA-Adapter parameters across sequential tasks, ensuring new task adaptations occupy parameter subspaces orthogonal to previous tasks' frozen parameter subspaces. Formally, the constraints for the \(t\)-th task are defined as:
\begin{equation}
\langle {\mathcal{W}}_2^{(t)}[:,i], \widetilde{\mathcal{W}}_2^{(s)}[:,j] \rangle = 0
,\ \forall i,\ j,\ s < t
\label{orth_constraint}
\end{equation}
The columns of \(\widetilde{\mathcal{W}}_2^{(t)}\) inherit directional properties from \(\mathcal{W}_2^{(t)}\), ensuring orthogonal relationships persist regardless of dynamic dimension activation patterns. These asymmetric orthogonality constraints enable simultaneous optimization of dynamic bottleneck dimension adaptation and historical knowledge preservation. We detail the analysis in Appendix~\ref{Appendix:why-orth-left}. To formalize this approach, we incorporate an orthogonality regularization term into the optimization objective. Specifically, the pairwise orthogonality loss between current task \(t\) and each historical task \(s<t\) is quantified as: 
\begin{equation}
\mathcal{L}_{\mathrm{orth}}^{(s,t)} = \sum_{i,j} \left\langle {\mathcal{W}}_2^{(t)}[:,i], \widetilde{\mathcal{W}}_2^{(s)}[:,j] \right\rangle^2
\end{equation}
Minimizing the loss term \(\mathcal{L}_{\mathrm{orth}}^{(s,t)}\) drives the inner product \(\langle {\mathcal{W}}_2^{(t)}[:,i], \widetilde{\mathcal{W}}_2^{(s)}[:,j] \rangle\) toward zero, enforcing parameter subspace orthogonality. The complete training objective, integrating both task-specific performance term \(\mathcal{L}_{\mathrm{task}}^{(t)}\) and orthogonality constraints term \(\mathcal{L}_{\mathrm{orth}}^{(s,t)}\), is formulated as:
\begin{equation}
\mathcal{L}_{\mathrm{total}} = \mathcal{L}_{\mathrm{task}}^{(t)} + \lambda_{\mathrm{orth}} \cdot \sum_{s<t}\mathcal{L}_{\mathrm{orth}}^{(s,t)}
\end{equation}
where \(\lambda_{\mathrm{orth}}\) is a hyperparameter controlling the strength of orthogonal regularization.

\section{Experiments and Analysis}
\label{Experiments}

\subsection{Experimental Settings.}
\paragraph{Datasets.}
We evaluate OA-Adapter using three CL benchmarks for LLMs: 
1) \textbf{Standard CL Benchmark}~\citep{DBLP:conf/nips/ZhangZL15}: a continual learning benchmark comprising five text classification datasets: AG News, Amazon Reviews, Yelp Reviews, DBpedia, and Yahoo Answers. 
2) \textbf{Long Sequence Benchmark}~\citep{DBLP:conf/iclr/RazdaibiedinaMH23}: a continual learning benchmark of 15 classification datasets, including five tasks from the Standard CL Benchmark, four from the GLUE benchmark (MNLI, QQP, RTE, SST2)~\citep{DBLP:conf/iclr/WangSMHLB19}, five from the SuperGLUE benchmark (WiC, CB, COPA, MultiRC, BoolQA)~\citep{wang2019superglue}, and the IMDB movie reviews dataset~\citep{DBLP:conf/acl/MaasDPHNP11}. 
3) \textbf{SuperNI Benchmark}~\citep{DBLP:conf/emnlp/WangMAKMNADASPK22}: a benchmark of diverse NLP tasks with expert-written instructions, enabling rigorous benchmarking of practical continual learning settings for LLMs. Specifically, we focus on five task categories: dialogue generation, information extraction, question answering, summarization, and sentiment analysis. From each category, three tasks are selected, resulting in a sequence of 15 tasks. 
Following~\citet{DBLP:conf/iclr/RazdaibiedinaMH23, DBLP:conf/acl/ZhaoWHZQZYXC24}, for all benchmarks we randomly select 1,000 samples per class for training.
The task details and training sequences of tasks used in our experiments are provided in Appendix~\ref{Appendix:datasets and orders}. 

\paragraph{Metrics.} 
Let \(a_{i,j}\) denote the test accuracy on the \(i\)-th task after training on the \(j\)-th task, the metrics for evaluating are:
1) \textbf{Average Accuracy(AA)}~\citep{DBLP:conf/eccv/ChaudhryDAT18} over all tasks after completing training on the final task, i.e., \(\text{AA}_T=\frac{1}{T} \sum_{i=1}^{T} a_{i,T}\).
2) \textbf{Forward Transfer (FWT)}~\citep{DBLP:conf/nips/Lopez-PazR17} which measures how much knowledge of previous tasks transfers to a new task, i.e., \(\text{FWT}_T = \frac{1}{T} \sum_{t=1}^{T} \left( a_{t,t} - a_{0,t} \right)\), where $a_{0,t}$ refers to the performance of training task $t$ individually.  
3) \textbf{Backward Transfer (BWT)}~\citep{DBLP:journals/corr/abs-2211-12701} which measures how much the learning of subsequent tasks influences the performance of a learned task, i.e.,\(\text{BWT}_T = \frac{1}{T-1} \sum_{t=1}^{T-1} \left( a_{T,t} - a_{t,t} \right)\)

\paragraph{Baselines.}
We compare our method against various CL baseline approaches, including:
\textbf{Replay} finetunes the whole model with a memory buffer, and replay samples from old tasks when learning new tasks to avoid forgetting.
\textbf{L2P~\citep{DBLP:conf/cvpr/0002ZL0SRSPDP22}} uses the input to dynamically select and update prompts from the prompt pool in an instance-wise fashion.
\textbf{LFPT5~\citep{qin2021lfpt5}} continuously train a soft prompt that simultaneously learns to solve the tasks and generate training samples, which are subsequently used in experience replay.
\textbf{O-LoRA~\citep{DBLP:conf/emnlp/WangCGXBZZGH23}} train new LoRA parameters on a sequential series of tasks in orthogonal subspace while fixing the LoRA matrices of previous tasks.
\textbf{ProgPrompt~\citep{DBLP:conf/iclr/RazdaibiedinaMH23}} adopts a task-specific soft prompt for each distinct task, sequentially appending it to prior learned prompts. In essence, it trains individual models per task, leveraging the task ID to select the appropriate model during inference.
In our continual learning baseline selection, we specifically focused on methods that could be reliably reproduced to ensure fair comparison. Furthermore, to guarantee the authenticity and consistency of our experimental results, we reproduced all baseline methods in our infrastructure.

\subsection{Main Results}
Our experiments employ both the encoder-decoder T5 model~\citep{DBLP:journals/jmlr/RaffelSRLNMZLL20} and the decoder-only LLaMA-7B model~\citep{DBLP:journals/corr/abs-2302-13971}, consistent with baselines in CL for NLP. Following previous works~\citep{qin2021lfpt5,DBLP:conf/emnlp/WangCGXBZZGH23}, we report
the results of three independent runs with different task orders on each CL benchmark, in Table~\ref{tab:baseline}. All experimental results are reported as the average of three runs. For more detailed settings, refer to Appendix~\ref{Appendix:implementation detail}. 

\begin{table}[!t]
\begin{minipage}{1.0\textwidth}
    \centering
    \caption{Testing performance on three benchmarks with T5-large.}
    \resizebox{\textwidth}{!}{
        \begin{tabular}{l|>{\centering\arraybackslash}p{0.8cm}>{\centering\arraybackslash}p{0.8cm}>{\centering\arraybackslash}p{0.8cm}|>{\centering\arraybackslash}p{1cm}>{\centering\arraybackslash}p{1cm}>{\centering\arraybackslash}p{1cm}|>{\centering\arraybackslash}p{0.8cm}>{\centering\arraybackslash}p{0.8cm}>{\centering\arraybackslash}p{0.8cm}}
\toprule
& \multicolumn{3}{c|}{Standard CL Benchmark} & \multicolumn{3}{c|}{Long Sequence Benchmark} & \multicolumn{3}{c}{SuperNI Benchmark} \\
            & AA            & FWT           & BWT           & AA            & FWT           & BWT           & AA            & FWT       & BWT           \\
\midrule
Replay      & 57.8          & -8.4          & -13.4         & 54.2          & 1.2           & -12.2         & 20.5          & -1.4                  & -15.8         \\
L2P         & 60.7          & -3.1          & -11.2         & 56.1          & 1.4           & -16.6         & 12.7          & -19.1                 & -8.0          \\
LFPT5       & 72.6          & \textbf{-1.9} & -8.3          & 68.6          & \textbf{2.7}  & -12.8         & 26.7          & \textbf{-0.5} & -14.5         \\
O-LoRA      & 75.3          & -3.6          & -9.1          & 68.7          & -6.2          & -4.1          & 25.9          & -7.8                  & -24.6         \\
ProgPrompt  & 75.1          & -2.3          & -8.1          & 63.2          & -2.3          & -6.7          & 23.3          & -3.3                  & -13.2         \\
OA-Adapter  & \textbf{76.0}  & -2.7         & \textbf{-7.5} & \textbf{69.2} & -4.4          & \textbf{-3.2} & \textbf{29.3} & -5.2                  & \textbf{-6.0} \\
\bottomrule
\end{tabular}

    }
    \label{tab:baseline}
\end{minipage}
\end{table}

\paragraph{Results on Continual Learning Benchmarks.}
As illustrated in Table~\ref{tab:baseline}, across all task orders of the standard CL benchmark, OA-Adapter consistently surpasses previous methods by a significant margin. 
On the other two benchmarks comprising 15 tasks, OA-Adapter consistently outperforms previous methods. 
Notably, on the SuperNI dataset, which is primarily designed for generative tasks, CL methods still perform poorly, underscoring that continual learning for complex tasks remains a significant challenge.
To disentangle the roles of orthogonal constraints and adaptive budget, we compare the full OA-Adapter with two variants: a no-orthogonality variant that removes the orthogonal parameter subspace constraint, and a fixed budget variant that disables dynamic bottleneck dimension adaptation. As shown in Table~\ref{tab:ablation}, removing orthogonality leads to significant degradation, underscoring its importance in mitigating catastrophic forgetting; disabling adaptation also reduces performance, indicating that adaptive capacity allocation provides additional gains.

\paragraph{Parameter Efficiency Analysis.}
As established in Section~\ref{Section:Dynamic Bottleneck Dimension Adaptation}, OA-Adapter leverages dynamic parameter budget allocation to achieve enhanced parameter efficiency. We compare the parameter utilization between OA-Adapter and O-LoRA across various initial budget conditions, as illustrated in Table~\ref{tab:parameter_effiency}. For OA-Adapter, budget allocation represents bottleneck dimension distribution, while for O-LoRA, it determines the module's intrinsic rank. Remarkably, OA-Adapter achieves superior performance while using \(46.6\%\) to \(58.5\%\) fewer parameters compared to O-LoRA's fixed budget approach. Moreover, OA-Adapter maintains consistent performance excellence across all tested initial budget settings, demonstrating its robust dynamic allocation capabilities. These results highlight our method's ability to effectively adapt parameter budget allocation according to task-specific requirements, providing substantial efficiency benefits over static parameter budget allocation approaches. 

\begin{table}[htbp]
\centering
\begin{minipage}{0.49\textwidth}
    \centering
    \caption{Ablation study evaluating the effects of orthogonality and budget adaptation.}
    \resizebox{\textwidth}{!}{
        \begin{tabular}{cccc}
\toprule

 \multirow{2}{*}{\textbf{Method}}   & \multicolumn{3}{c}{\textbf{Dataset}} \\
          & Standard & Long  & SuperNI \\ \midrule
OA-Adapter  & 76.0                    & 69.2                    & 29.3  
        \\ 
- Orthogonality & 57.1                  & 52.8                    & 13.7              \\
- Budget Adaptation   & 73.3                  & 65.6                    & 24.2              \\

\bottomrule
\end{tabular}
    }
    \label{tab:ablation}
\end{minipage}
\hfill
\begin{minipage}{0.50\textwidth}
    \centering
    \caption{Comparisons of parameter efficiency between OA-Adapter and O-LoRA.}
    \resizebox{\textwidth}{!}{
        \begin{tabular}{cccc}
\toprule
 \textbf{Method}     & \textbf{Budget Allocation} & \textbf{Params} & \textbf{AA}  \\
\midrule
O-LoRA     & {16→16} & 4.72M      & 75.3            \\
OA-Adapter & {16→9.95} & {1.96M}{\color{green!60!black}{\scriptsize -58.5\%}} & {76.0}{\color{red!90!black}{\scriptsize +0.7}} \\
\midrule
O-LoRA     & {8→8} & 2.36M      & 74.5            \\
OA-Adapter & {8→6.05} & {1.18M}{\color{green!60!black}{\scriptsize -50.0\%}} & {74.7}{\color{red!90!black}{\scriptsize +0.2}} \\
\midrule
O-LoRA     & {4→4} & 1.18M      & 73.8            \\
OA-Adapter & {4→3.18} & {0.63M}{\color{green!60!black}{\scriptsize -46.6\%}} & {74.1}{\color{red!90!black}{\scriptsize +0.3}} \\
\bottomrule                       
\end{tabular}

    }
    \label{tab:parameter_effiency}
\end{minipage}
\end{table}

\begin{figure}[htbp]
\centering
\includegraphics[width=0.8 \linewidth]{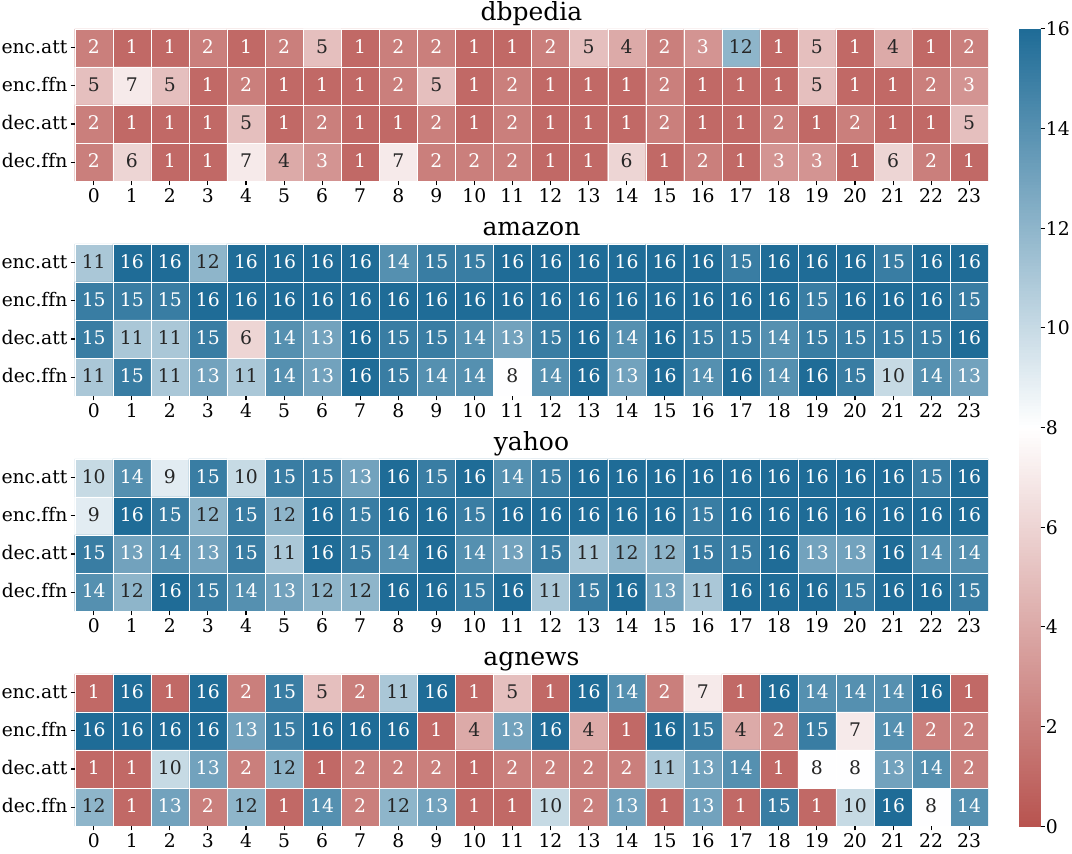}
\caption{Final dimensions after sequential training following Order-1 with OA-Adapter across four text classification datasets (i.e., DBpedia, Amazon, Yahoo, AG News). The X-axis is the index of T5-large layers, and the Y-axis indicates different layers OA-Adapter applies to.}
\label{fig:sparse_rank}
\end{figure}

\paragraph{Heterogeneous Budget Requirements Across Tasks and Layers.}
\label{Exp:final dimension}
Intuitively, adapting to individual downstream datasets requires varying parameter budgets across different tasks and layers. To validate this, we analyze budget allocation patterns in CL scenarios, as shown in Figure~\ref{fig:sparse_rank}. Our results reveal heterogeneous budget requirements across tasks and layer positions, confirming that optimal parameter allocation cannot follow uniform rules but demands task-specific consideration.
Notably, in CL scenarios, the parameter matrices for the initial task exhibit significantly higher sparsity compared to subsequent tasks. This pattern supports our hypothesis that initial tasks primarily leverage capabilities inherent in the pretrained model, while later tasks must additionally preserve knowledge from preceding tasks, necessitating more complex parameter spaces. Comprehensive analysis is provided in Appendix~\ref{Appendix:final dimension}. These findings validate the necessity of adaptive budget allocation for CL based on the characteristics of layer, task and training sequence.

\subsection{Discussions and Analysis}
\captionsetup[figure]{skip=3pt} 
\begin{wrapfigure}{r}{0.55\linewidth}
\vspace{-30pt} 
\centering
\includegraphics[width=\linewidth]{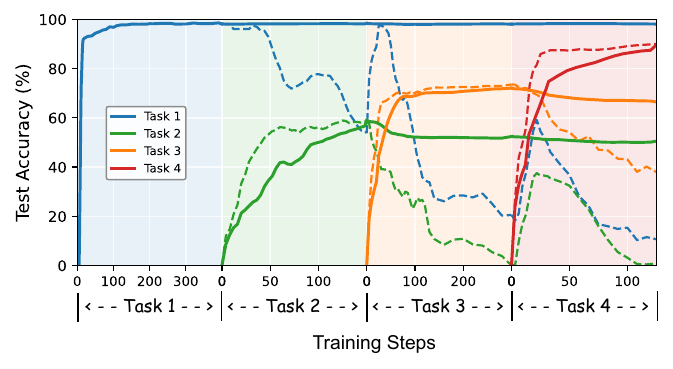}
\caption{Occurrence and mitigation of catastrophic forgetting during sequential training following Order-1 across multiple tasks.}
\label{fig:orthogonality}
\vspace{-5pt} 
\end{wrapfigure}

\paragraph{Occurrence and Mitigation of Catastrophic Forgetting.}

\label{Exp:catastrophic forgetting}
As shown in Figure~\ref{fig:orthogonality}, we demonstrate the occurrence and mitigation of catastrophic forgetting on the standard CL benchmark. Solid and dashed lines denote models with and without orthogonal parameter-subspace constraints, and each color corresponds to a distinct task. Shaded background bands mark the training phase of each task, and the X-axis is scaled proportionally to training steps.
We observe a significant decline of each task after their training phase without orthogonal constraints. Performance on Task 2 drops to nearly zero by the end of the subsequent tasks, while Task 1 and Task 3 also decrease substantially. 
In contrast, the performance with orthogonal constraints remains largely maintained through Task 4, with the most severe degradation limited to only \(14\%\) performance loss on Task 2. 
These results demonstrate that severe forgetting phenomena occur during multi-task training and orthogonal parameter subspace constraints can effectively mitigate it. Similar trends are consistently observed under other task orders, as detailed in Appendix~\ref{Appendix:orthogonal ablation}.
Additionally, as expected, each task achieves higher performance more rapidly during its corresponding training phase without orthogonal constraints, though final performance is not substantially higher than with orthogonal constraints. This suggests that the model has greater flexibility in parameter update directions when not constrained to preserve previous knowledge, and orthogonal subspace still maintain sufficient capacity for subsequent tasks.
Interestingly, tasks exhibit brief performance recovery at the beginning of subsequent training phases before further decline. Performance on Task 1 recovers to nearly \(100\%\) at the start of Task 3 and to approximately \(60\%\) during Task 4. 
Although Task 2 drops to near-zero by the end of Task 3, it improves rapidly early in Task 4 training. 
This behavior resembles human memory, where disused knowledge can be reactivated with minimal effort. This suggests that catastrophic forgetting may only superficially mask a latent capacity, and knowledge representations remain partially preserved despite significant performance degradation.


\paragraph{Budget Adaptation Mechanism Analysis.}
To assess the impact of our threshold strategy, we compare two policies: (a) a fixed, non-learnable threshold and (b) our proposed learnable threshold that adapts across different layers and tasks during training, as described in Section~\ref{Section:Dynamic Bottleneck Dimension Adaptation}. These strategies are assessed using the T5-large model across three task orders on the standard continual learning benchmark. The results, depicted in Table~\ref{tab:threshold}, reveal that the dynamic threshold consistently demonstrates superior performance compared to the fixed threshold. This confirms that the dynamic threshold mechanism enables bidirectional adjustment of budget allocation without introducing complex mechanisms, thereby enhancing more flexibility in the optimization process.
\begin{table}[htbp]
\centering
\begin{minipage}{0.53\textwidth}
    \centering
    \caption{Comparisons of threshold strategies.}
    \resizebox{\textwidth}{!}{
        \begin{tabular}{cccccc}
\toprule
\multicolumn{2}{c}{\textbf{Threshold}}     & \multicolumn{4}{c}{\textbf{order}}                  \\
\textbf{Initial Value} & \textbf{Strategy} & \textbf{1} & \textbf{2} & \textbf{3} & \textbf{avg} \\
\midrule                  
\multirow{2}{*}{1e-3} & fixed   & 71.5  & 71.1  & 71.4  & 71.3  \\
& dynamic & \textbf{73.0}  & \textbf{76.4}  & \textbf{74.6}  & \textbf{74.7}  \\
\midrule
\multirow{2}{*}{1e-4} & fixed   & 73.7  & 73.1  & 71.4  & 72.7  \\
& dynamic & \textbf{75.7}  & \textbf{75.6}  & \textbf{75.1}  & \textbf{75.5}  \\
\midrule
\multirow{2}{*}{1e-5} & fixed   & 72.8  & 73.5  & 72.3  & 72.9  \\
& dynamic & \textbf{74.8}  & \textbf{74.9}  & \textbf{73.9}  & \textbf{74.5}  \\
\bottomrule
\end{tabular}
    }
    \label{tab:threshold}
\end{minipage}
\hfill
\begin{minipage}{0.46\textwidth}
    \centering
    \caption{Comparisons of pre-trained models.}
    \resizebox{\textwidth}{!}{
        \begin{tabular}{cccccc}
\toprule
\multirow{2}{*}{\textbf{Model}} & \multirow{2}{*}{\textbf{Method}}  & \multicolumn{4}{c}{\textbf{order}}                  \\
 & & \textbf{1} & \textbf{2} & \textbf{3} & \textbf{avg} \\
\midrule
\multirow{2}{*}{T5-base} & O-LoRA   & 73.9       & \textbf{74.8}       & 74.1       & 74.3         \\
& OA-Adapter & \textbf{74.7}       & \textbf{74.8}       & \textbf{74.5}        & \textbf{74.7}          \\ 
\midrule                  
\multirow{2}{*}{T5-large} & O-LoRA   & 75.0       & 75.4       & 75.6       & 75.3         \\
& OA-Adapter & \textbf{75.7}       & \textbf{76.2}       & \textbf{76.1}       & \textbf{76.0}         \\ 
\midrule
\multirow{2}{*}{T5-XL} & O-LoRA   & 77.9  & \textbf{78.5}  & 77.4  & 77.9  \\
& OA-Adapter & \textbf{78.0}  & 78.2  & \textbf{77.9}  & \textbf{78.0}  \\
\midrule
\multirow{2}{*}{Llama-7B} & O-LoRA   & 76.2  & 76.3  & 75.7  & 76.1  \\
& OA-Adapter & \textbf{76.4}  & \textbf{76.8}  & \textbf{76.7}  & \textbf{76.6}  \\
\bottomrule
\end{tabular}
    }
    \label{tab:model}
\end{minipage}
\end{table}

\paragraph{Pre-trained Model Analysis.}
We investigate the performance of OA-Adapter and O-LoRA across varying models (T5-base, T5-large, T5-XL, Llama-7B) using the standard continual learning benchmark. The results, depicted in Table~\ref{tab:model}, reveal that OA-Adapter's average accuracy consistently improves, which suggests that OA-Adapter effectively leverages the increased representational capacity of larger models. The consistent superiority across different models indicates that OA-Adapter's mechanism provides effective protection against catastrophic forgetting while enabling precise task-specific optimization, regardless of the underlying model architecture's complexity and scales. 

\paragraph{Compatibility with Replay.}
To assess compatibility with replay-based strategies, we evaluate a hybrid variant OA-Adapter+Replay and compare it with the strong replay baseline SAPT~\citep{DBLP:conf/acl/ZhaoWHZQZYXC24}. The hybrid consistently yields additional gains while preserving low forgetting, indicating that our method complements replay. Full results and settings are provided in Appendix~\ref{Appendix: replay}.

\section{Conclusion}
\label{Conclusion}
In this paper, we introduce OA-Adapter, a novel parameter-efficient approach for continual learning in LLMs that considers both dynamic budget adaptation and orthogonal subspace learning in an end-to-end training stage. Our comprehensive experiments demonstrate OA-Adapter's consistent superiority over existing methods across multiple benchmarks while using significantly fewer parameters. The observed heterogeneity in optimal parameter allocation across tasks and layers validates the necessity of our budget-adaptive approach. As the first work to integrate budget adaptation into parameter-efficient fine-tuning for continual learning in LLMs, OA-Adapter establishes a new paradigm that jointly optimizes parameter budget allocation and knowledge preservation. This advancement paves the way for efficient and effective adaptation of LLMs to evolving real-world. 

\section{Reproducibility Statement}
It is important to note that the work presented in this paper is reproducible. 
To ensure the reproducibility of our results, we have made several efforts, which we summarize below.
First, the experiments conducted in this paper, including the models and datasets used, are described in detail in Section~\ref{Experiments}. 
For more specific information on the datasets, we refer readers to Appendix~\ref{Appendix:datasets and orders}, which provides the complete details. 
The implementation details of the models, including hyperparameter settings and optimization procedures, are provided in Appendix~\ref{Appendix:implementation detail}.
We have submitted the source code as supplementary material, which includes all implementations necessary to reproduce the experiments. 
After acceptance, the code will be publicly available on GitHub for the research community. This code includes the OA-Adapter model, training routines, and configuration files.
By providing these detailed resources, we aim to ensure that our work can be reproduced accurately. 
Furthermore, we encourage others to conduct further exploration and research based on our work.

\bibliography{main}
\bibliographystyle{iclr2026_conference}

\appendix
\section{Appendix}

\subsection{LLM Usage Declaration}
Large language models (e.g., ChatGPT) were used solely for language editing and formatting. 
They did not contribute to the conception, design, implementation, analysis, data generation or labeling, 
or evaluation of the methods and results. All technical content and claims were authored and verified by the authors, 
and no personal, proprietary, or sensitive data were shared with LLM services.

\subsection{Details About Orthogonality Constraints}
\label{Appendix:why-orth-left}

In Equation (\ref{parameter}) and (\ref{orth_constraint}), OA-Adapter choose \( \mathcal{W}_2^{(k)} \cdot \Gamma^{(k)} \) rather than \( \Gamma^{(k)} \cdot \mathcal{W}_1^{(k)} \) to enforce the orthogonal relationships. Here’s why:

First, consider an OA-Adapter layer without the gating matrix \(\Gamma^{(k)}\). Let \( W_1 \in \mathbb{R}^{r \times d}\) and \( W_2 \in \mathbb{R}^{d \times r} \) be the projection matrices for the down-projection and up-projection, respectively. For an input vector \(x \in \mathbb{R}^d\), the adapter’s output is:\(y = W_2 \cdot W_1 \cdot x\). Because \(W_2\) is composed of column vectors \(w_{2,1},\dots,w_{2,r}\), this can be rewritten as \(y=\sum_{i=1}^r \alpha_i w_{2,i}\), where the scalars \(\alpha_i\) are the \(i\)-th components of the intermediate vector 
\(W_1 \cdot x\). Hence the output \(y\) is a linear combination of the columns of \(W_2\).And the \(W_1\) merely supplies the mixing coefficients.

Furthermore, in the full OA-Adapter, we introduce a diagonal gate \(\Gamma^{(k)}\), for each task \(k\) to selectively activate a subset of the basis directions, and then we impose the orthogonality constraint on the product \(W_2^{(k)}\Gamma^{(k)}\). This choice is deliberate: \(W_2^{(k)}\Gamma^{(k)}\) directly determines the directions injected into the transformer’s hidden state, whereas \(W_1^{(k)}\) only controls how strongly each direction is used. If orthogonality were enforced on \(\Gamma^{(k)}W_1^{(k)}\) instead, the subsequent multiplication by \(W_2^{(k)}\) would generally break that orthogonality, allowing task interference to resurface. Constraining \(W_2^{(k)}\Gamma^{(k)}\) keeps the basis directions themselves orthogonal across tasks, which is precisely what reduces interference in continual learning.

\subsection{Datasets and Task Orders.}

\label{Appendix:datasets and orders}
Table~\ref{tab:long_detail} and ~\ref{tab:superni_detail} show details of the datasets we used for our experiments, along with their evaluation metrics. Overall, we used datasets from the Standard CL benchmark~\cite{DBLP:conf/nips/ZhangZL15}, GLUE~\cite{DBLP:conf/iclr/WangSMHLB19}, and SuperGLUE~\cite{wang2019superglue} benchmarks, and added the IMDB movie reviews dataset~\cite{DBLP:conf/acl/MaasDPHNP11}, following~\cite{DBLP:conf/iclr/RazdaibiedinaMH23}. Additionally, we incorporated the SuperNI Benchmark~\cite{DBLP:conf/emnlp/WangMAKMNADASPK22}, a benchmark of diverse NLP tasks with expert-written instructions, for more rigorous evaluation of our model.
Furthermore, Table~\ref{tab:order_detail} shows details of task orders used in our CL experiments.

\begin{table}[htbp]
\caption{The details of 15 classification datasets in the Long Sequence Benchmark. NLI denotes natural language inference, QA denotes questions and answers task. First five tasks correspond to the standard CL benchmark, all other tasks are used in long-sequence experiments.}
\label{tab:long_detail}
\centering
\begin{tabular}{cllll}
\toprule
\textbf{Dataset} & \textbf{Category} & \textbf{Task} & \textbf{Domain} & \textbf{Metric} \\
\midrule
Yelp & CL Benchmark & Sentiment Analysis & Yelp Reviews & Accuracy \\
Amazon & CL Benchmark & Sentiment Analysis & Amazon Reviews & Accuracy \\
DBPedia & CL Benchmark & Topic Classification & Wikipedia & Accuracy \\
Yahoo & CL Benchmark & Topic Classification & Yahoo Q\&A & Accuracy \\
AG News & CL Benchmark & Topic Classification & News & Accuracy \\
MNLI & GLUE & NLI & Various & Accuracy \\
QQP & GLUE & Paraphrase Detection & Quora & Accuracy \\
RTE & GLUE & NLI & News, Wikipedia & Accuracy \\
SST-2 & GLUE & Sentiment Analysis & Movie Reviews & Accuracy \\
WIC & SuperGLUE & Word Sense Disambiguation & Lexical Databases & Accuracy \\
CB & SuperGLUE & NLI & Various & Accuracy \\
COPA & SuperGLUE & QA & Blogs, Encyclopedia & Accuracy \\
BoolQA & SuperGLUE & Boolean QA & Wikipedia & Accuracy \\
MultiRC & SuperGLUE & QA & Various & Accuracy \\
IMDB & SuperGLUE & Sentiment Analysis & Movie Reviews & Accuracy \\
\bottomrule
\end{tabular}
\end{table}

\begin{table}[htbp]
\caption{The details of 15 datasets in the SuperNI Benchmark.}
\label{tab:superni_detail}
\centering
\begin{tabular}{lll}
\toprule
\textbf{Dataset name}                              & \textbf{Task}          & \textbf{Metric} \\
\midrule
task639\_multi\_woz\_user\_utterance\_generation   & dialogue generation    & Rouge-L         \\
task1590\_diplomacy\_text\_generation              & dialogue generation    & Rouge-L         \\
task1729\_personachat\_generate\_next              & dialogue generation    & Rouge-L         \\
task181\_outcome\_extraction                       & information extraction & Rouge-L         \\
task748\_glucose\_reverse\_cause\_event\_detection & information extraction & Rouge-L         \\
task1510\_evalution\_relation\_extraction          & information extraction & Rouge-L         \\
task002\_quoref\_answer\_generation                & question answering     & Rouge-L         \\
task073\_commonsenseqa\_answer\_generation         & question answering     & Rouge-L         \\
task591\_sciq\_answer\_generation                  & question answering     & Rouge-L         \\
task511\_reddit\_tifu\_long\_text\_summarization   & summarization          & Rouge-L         \\
task1290\_xsum\_summarization                      & summarization          & Rouge-L         \\
task1572\_samsum\_summary                          & summarization          & Rouge-L         \\
task363\_sst2\_polarity\_classification            & sentiment analysis     & accuracy        \\
task875\_emotion\_classification                   & sentiment analysis     & accuracy        \\
task1687\_sentiment140\_classification             & sentiment analysis     & accuracy    \\
\bottomrule
\end{tabular}
\end{table}

\begin{table}[htbp]
\caption{Eight different task sequence orders utilized in our experiments. Orders 1-3 follow the standard continual learning benchmark as established by previous research, focusing on a more traditional task sequence. Orders 4-6 are customized for long-sequence experimentation, encompassing 15 tasks each and are structured according to the methodologies outlined in~\cite{DBLP:conf/iclr/RazdaibiedinaMH23}. Orders 7-8 correspond to the SuperNI benchmark.}
\label{tab:order_detail}
\centering
\begin{tabular}{ccc}
\toprule
\textbf{Order} & \textbf{Task Sequence} \\
\midrule
1 & dbpedia $\rightarrow$ amazon $\rightarrow$ yahoo $\rightarrow$ ag \\
2 & dbpedia $\rightarrow$ amazon $\rightarrow$ ag $\rightarrow$ yahoo \\
3 & yahoo $\rightarrow$ amazon $\rightarrow$ ag $\rightarrow$ dbpedia \\
\midrule
\multirow{2}{*}{4} & mnli $\rightarrow$ cb $\rightarrow$ wic $\rightarrow$ copa $\rightarrow$ qqp $\rightarrow$ boolqa $\rightarrow$ rte $\rightarrow$ imdb $\rightarrow$\\
 & yelp $\rightarrow$ amazon $\rightarrow$ sst-2 $\rightarrow$ dbpedia $\rightarrow$ ag $\rightarrow$ multirc $\rightarrow$ yahoo \\
\multirow{2}{*}{5} & multirc $\rightarrow$ boolqa $\rightarrow$ wic $\rightarrow$ mnli $\rightarrow$ cb $\rightarrow$ copa $\rightarrow$ qqp $\rightarrow$ rte $\rightarrow$\\
 & imdb $\rightarrow$ sst-2 $\rightarrow$ yelp $\rightarrow$ amazon $\rightarrow$ ag $\rightarrow$ dbpedia $\rightarrow$ yahoo \\
\multirow{2}{*}{6} & yelp $\rightarrow$ amazon $\rightarrow$ mnli $\rightarrow$ cb $\rightarrow$ copa $\rightarrow$ qqp $\rightarrow$ rte $\rightarrow$ imdb $\rightarrow$\\
 & sst-2 $\rightarrow$ dbpedia $\rightarrow$ ag $\rightarrow$ yahoo $\rightarrow$ multirc $\rightarrow$ boolqa $\rightarrow$ wic \\
\midrule
\multirow{3}{*}{7} & task1572 $\rightarrow$ task363 $\rightarrow$ task1290 $\rightarrow$ task181 $\rightarrow$ task002 $\rightarrow$  \\
                   & task1510 $\rightarrow$ task639 $\rightarrow$ task1729 $\rightarrow$ task073 $\rightarrow$ task1590 $\rightarrow$ \\
                   & task748 $\rightarrow$ task511 $\rightarrow$ task591 $\rightarrow$ task1687 $\rightarrow$ task875     \\
\multirow{3}{*}{8} & task748 $\rightarrow$ task073 $\rightarrow$ task1590 $\rightarrow$ task639 $\rightarrow$ task1572 $\rightarrow$  \\
                   & task1687 $\rightarrow$ task591 $\rightarrow$ task363 $\rightarrow$ task1510 $\rightarrow$ task1729 $\rightarrow$ \\
                   & task181 $\rightarrow$ task511 $\rightarrow$ task002 $\rightarrow$ task1290 $\rightarrow$ task875  \\
\bottomrule
\end{tabular}
\end{table}

\subsection{Implementation Detail.}
\label{Appendix:implementation detail}
All our experiments involving T5 models were performed on a server outfitted with four NVIDIA GeForce RTX 3090 GPUs, utilizing the DeepSpeed repository for implementation. Following previous studies~\cite{DBLP:conf/nips/dAutumeRKY19,DBLP:conf/nips/RaoVRPTH19}, for CL experiments, for each dataset we use the available validation set as a test set (since test data is not available) and hold out 500 samples from the train set to construct the validation set. For every sequence of tasks across different orders, we trained the models for one epoch using a batch size of 32 (8 per GPU), a dropout rate of 0.1, and no weight decay. Across all experiments, we primarily used Adapter modules with a bottleneck dimension of 16, and applied a sparsification threshold chosen from \(\{1e\text{-}3, 1e\text{-}4, 1e\text{-}5\}\). The learning rate was selected from \(\{5e\text{-}3, 3e\text{-}3, 1e\text{-}3, 5e\text{-}4\}\) depending on task characteristics. We applied an orthogonality regularization on the Adapter’s upsampling matrix with a coefficient \(\lambda_{orth} \in \{0.5, 1, 5\}\), and used an additional coefficient \(\lambda_2 \in \{0, 0.1, 0.5\}\) to scale the associated L2 loss term. For T5-Large, we report the average accuracy (AA) for Orders 1-3 in Table~\ref{tab:o_lamda}, as the primary metric for evaluating the effect of $\lambda_{\text{orth}}$. Unless otherwise specified, the reported main results use the $\lambda_{\text{orth}}$ selected by the best validation AA averaged across orders. This protocol balances knowledge retention and sparsity, and yields stable performance across long sequences with distribution shift.
To ensure experimental comparability and fair result comparison, we maintain consistency with O-LoRA~\cite{DBLP:conf/emnlp/WangCGXBZZGH23} by adopting instruction tuning as the training paradigm across all experiments for both our method and other baselines, as shown in Table~\ref{tab:instructions}. This approach offers dual advantages: it incorporates human expertise for efficient learning while enabling models to better capture underlying principles through explicit guidance, thereby enhancing generalization capabilities. The consistent instruction-based framework allows for direct performance comparisons while leveraging the benefits of natural language supervision.
\begin{table}[htbp]
\caption{Effect of the orthogonality coefficient $\lambda_{\text{orth}}$ on T5-large.}
\label{tab:o_lamda}
\centering
\begin{tabular}{ccccc}
\toprule
 $\lambda_{\text{orth}}$   & Order 1 & Order 2 & Order 3 & Average \\
 \midrule
0.5 & 74.3    & 75.3    & 74.5    & 74.7    \\
1   & 73.4    & 74.8    & 75.1    & 74.4    \\
2   & 73.5    & 74.9    & 73.2    & 73.9    \\
3   & 74.2    & 73.9    & 74.1    & 74.1    \\
4   & 74.7    & 76.4    & 73.6    & 74.9    \\
5   & 75.8    & 75.6    & 74.6    & 75.3   \\
\bottomrule
\end{tabular}
\end{table}

\begin{table}[htbp]
\caption{Instructions for different tasks.}
\label{tab:instructions}
\centering
\begin{tabular}{cl}
\toprule
\textbf{Task} & \textbf{Prompts} \\
\midrule
\multirow{2}{*}{NLI} & What is the logical relationship between the "sentence 1" and the "sentence 2"? \\
 &Choose one from the option. \\
\midrule
\multirow{2}{*}{QQP} & Whether the "first sentence" and the "second sentence" have the same meaning? \\
 &Choose one from the option. \\
\midrule
SC & What is the sentiment of the following paragraph? Choose one from the option. \\
\midrule
TC & What is the topic of the following paragraph? Choose one from the option. \\
\midrule
\multirow{2}{*}{BoolQA} & According to the following passage, is the question true or false? \\
 & Choose one from the option. \\
\midrule
\multirow{2}{*}{MultiRC} & According to the following passage and question, is the candidate answer true \\
 & or false? Choose one from the option. \\
\midrule
\multirow{2}{*}{WiC} & Given a word and two sentences, whether the word is used with the same sense \\
 & in both sentence? Choose one from the option. \\
\bottomrule
\end{tabular}
\end{table}

\subsection{Heterogeneous Budget Requirements Across Tasks and Layers.}
\label{Appendix:final dimension}
\begin{figure}[htbp]
\centering
\includegraphics[width=0.8 \linewidth]{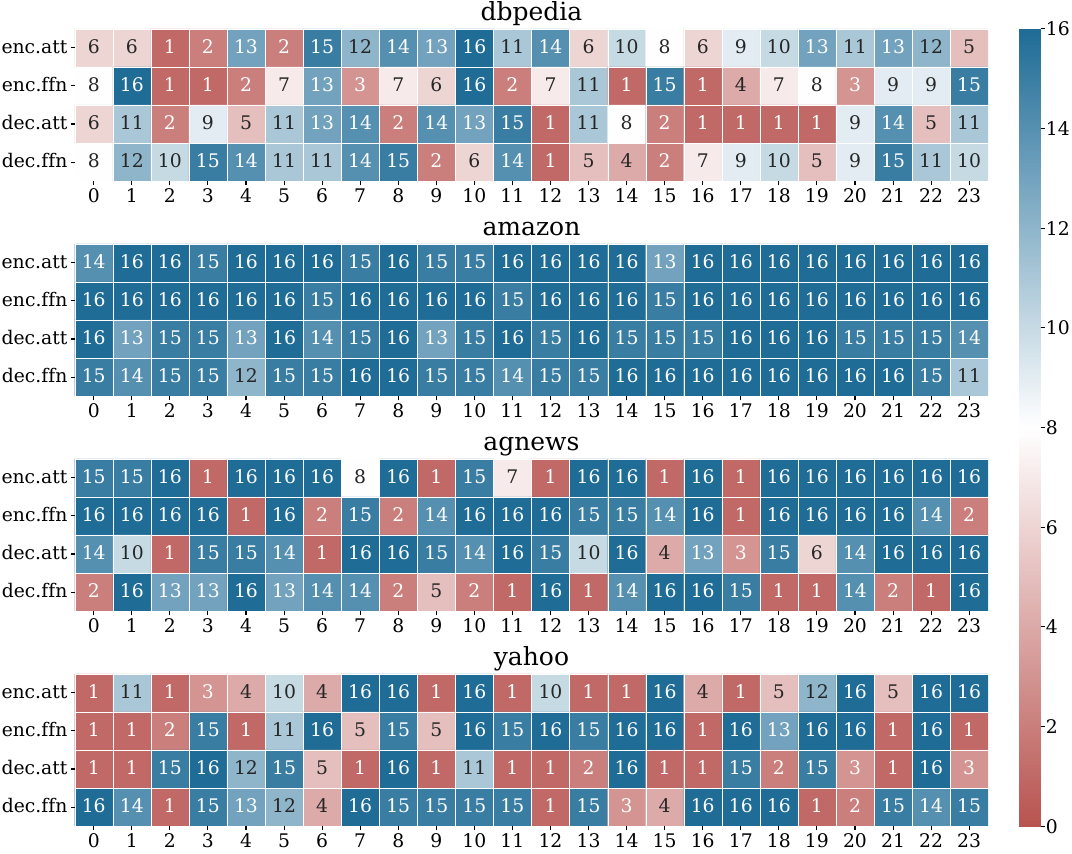}
\caption{Final dimensions after sequential training following Order-2 with OA-Adapter.}
\label{fig:sparse_rank_2}
\end{figure}

\begin{figure}[htbp]
\centering
\includegraphics[width=0.8 \linewidth]{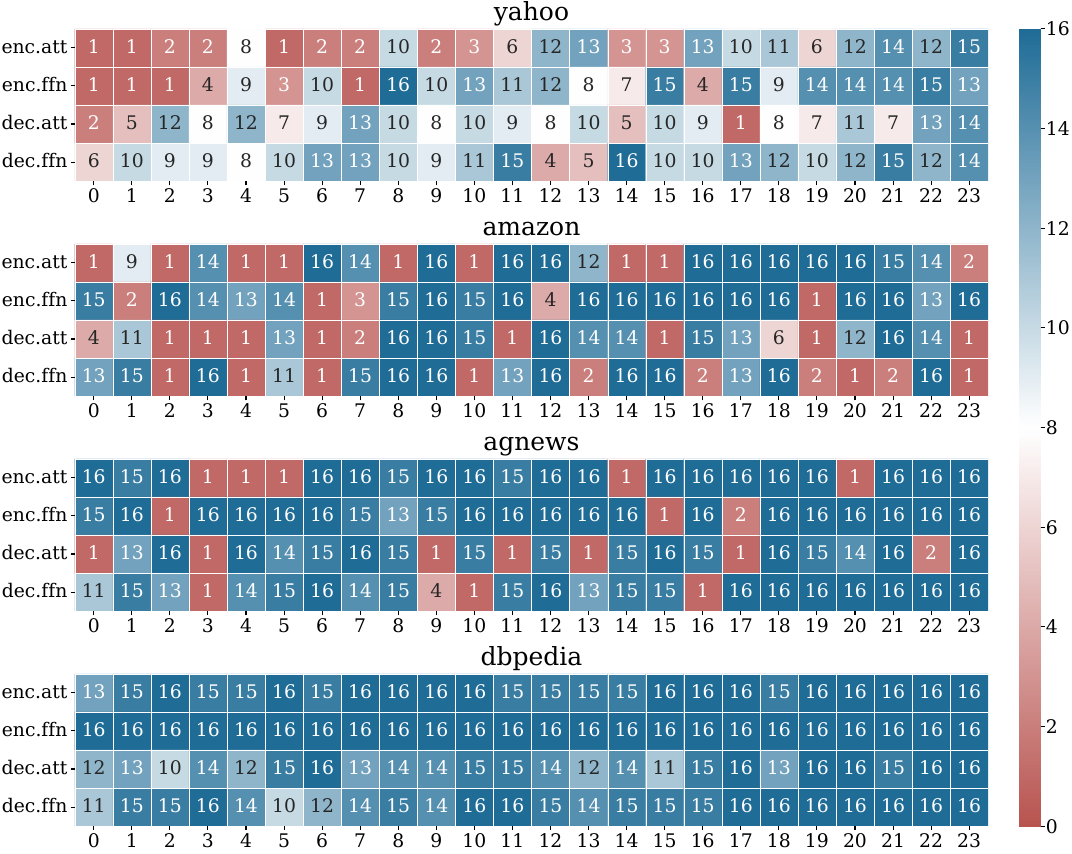}
\caption{Final dimensions after sequential training following Order-3 with OA-Adapter.}
\label{fig:sparse_rank_3}
\end{figure}

As discussed in Section~\ref{Exp:final dimension}, we extend our analysis to investigate budget allocation in the context of continual learning. Here, we further present results under other task orders, as illustrated in Figure~\ref{fig:sparse_rank_2} and~\ref{fig:sparse_rank_3}. These findings corroborate our analysis in Section~\ref{Exp:final dimension}: (a) The relationship between performance and parameter budget does not follow constant rules but rather necessitates case-specific consideration. (b) Within continual learning scenarios, the first task primarily focuses on acquiring capabilities built upon the pretrained model, whereas subsequent tasks must additionally preserve knowledge from preceding tasks, thus requiring more nuanced fine-tuning. This further substantiates our analysis that different tasks in the CL scenarios require varying budgets, and that allocating budgets according to training sequence and task characteristics is both necessary and justified. 


\subsection{Occurrence and Mitigation of Catastrophic forgetting.}
\label{Appendix:orthogonal ablation}
\begin{figure}[htbp] 
\centering
\includegraphics[width=1.0 \linewidth]{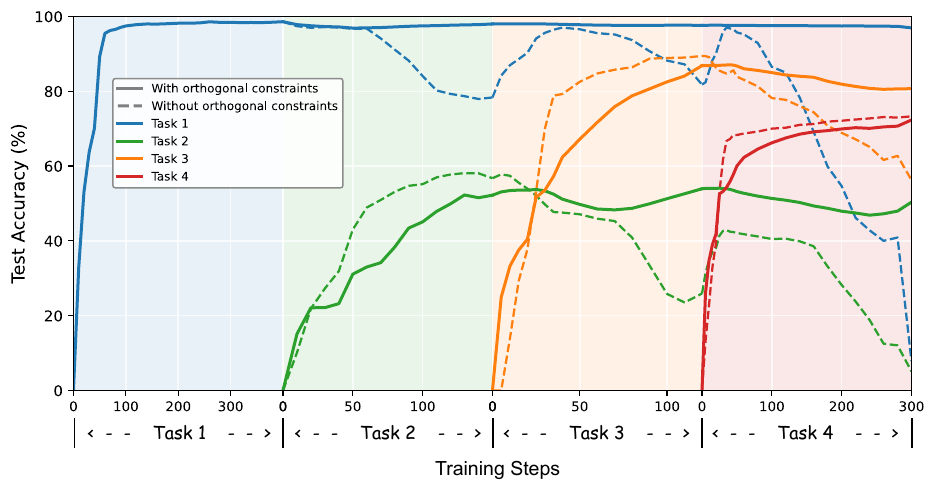}
\caption{Occurrence and mitigation of catastrophic forgetting during sequential training following Order-2 across multiple tasks.}
\label{fig:orthogonality_2}
\end{figure}

\begin{figure}[htbp] 
\centering
\includegraphics[width=1.0 \linewidth]{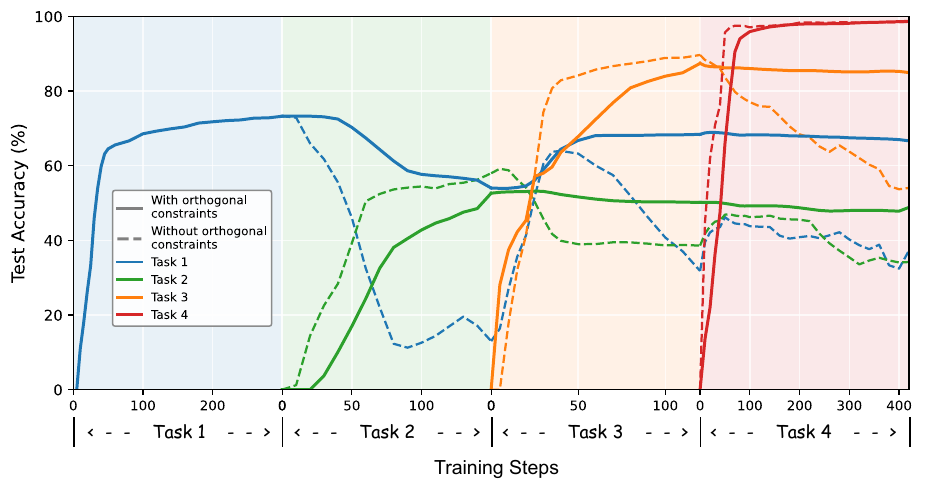}
\caption{Occurrence and mitigation of catastrophic forgetting during sequential training following Order-3 across multiple tasks.}
\label{fig:orthogonality_3}
\end{figure}

To further validate the effectiveness and consistency of orthogonal parameter subspace constraints, we conduct sequential training following Order-2 and Order-3, with results illustrated in Figure~\ref{fig:orthogonality_2} and~\ref{fig:orthogonality_3}, respectively. Consistent with the results reported in Section~\ref{Exp:catastrophic forgetting}, we observe severe catastrophic forgetting in the absence of orthogonal constraints, especially for earlier tasks. In contrast, models trained with orthogonal parameter subspace constraints are able to preserve performance across all tasks to a much greater extent. Notably, although the specific tasks affected most by forgetting vary depending on the task order, the general trend holds: orthogonal constraints provide consistent mitigation of forgetting regardless of task permutation. These results reinforce our earlier findings and highlight the robustness of orthogonal subspace regularization as a general mechanism for alleviating forgetting in continual learning scenarios.


\subsection{Compatibility with Replay.}
\label{Appendix: replay}
In this appendix, we discuss the compatibility of our method, OA-Adapter, with replay-based strategies. We compare this hybrid with SAPT~\citep{DBLP:conf/acl/ZhaoWHZQZYXC24}, a strong baseline in the replay-based continual learning setting.

Our experiments, conducted on the SuperNI Benchmark, show that the OA-Adapter+Replay hybrid consistently achieves additional gains in performance while maintaining low forgetting compared to the original OA-Adapter. Specifically, the average performance of our method is 48.3, outperforming SAPT+Replay’s 44.2 on the SuperNI dataset. This indicates that our method can effectively complement replay strategies and improve performance.

Although OA-Adapter+Replay achieves lower performance than SAPT+ARM (50.8), we note that our approach still demonstrates better compatibility with replay-based strategies, offering a more efficient and scalable solution.

However, it is important to note that replay-based methods come with significant overhead, either requiring the storage of large amounts of task data or generating synthetic data for retraining. Both approaches introduce challenges, including storage limitations and potential privacy concerns due to data retention. Additionally, replay-based methods often incur increased computational costs, as they necessitate continuous retraining with stored or generated data.

Given these challenges, while our method demonstrates compatibility with replay-based strategies and provides notable improvements, we chose not to incorporate replay into our final approach due to the associated resource-intensive nature and practical limitations. Instead, we focus on efficient, memory-conscious strategies that can be more easily scaled in real-world applications without the need for large-scale data storage or generation.

\subsection{Compute and Memory Cost Analysis}
\label{Appendix:cost-analysis}

\paragraph{Training cost and GPU memory.}
We benchmark the training cost of OA-Adapter against O-LoRA under Task Order~1 on the Standard CL benchmark. OA-Adapter demonstrated slightly lower GPU memory usage, averaging 68.9\% compared to 70.4\% for O-LoRA. OA-Adapter achieved a significantly faster average training time per step: 0.357 seconds versus 0.582 seconds for O-LoRA, suggesting a clear advantage in training efficiency.

\paragraph{Overhead from orthogonality regularization.}
We further isolate the per-step time consumed by orthogonality computations for OA-Adapter (bottleneck dimension 24) on the Standard CL benchmark, as shown in Table~\ref{tab:ortho_overhead}. The per-step overhead grows with the number of previously seen tasks, resulting in an average of 0.0179\,s per step across the first four tasks, i.e., a 4.6\% relative overhead compared to the total per-step time of 0.357\,s. As expected from the pairwise constraints with prior tasks, this cost grows approximately linearly with the task sequence length; extrapolating to a 15-task sequence yields an estimated proportion of 14.7\% of total training time devoted to orthogonality operations.

\begin{table}[htbp]
\centering
\caption{Per-step time spent on orthogonality computations for OA-Adapter.}
\label{tab:ortho_overhead}
\begin{tabular}{lccccc}
\toprule
Task & 1 & 2 & 3 & 4 &avg\\
\midrule
Time (s) & 0.0061 & 0.0130 & 0.0199 & 0.0269 & 0.0179 \\
\bottomrule
\end{tabular}
\end{table}


\end{document}